\documentclass[sigconf,nonacm]{acmart}
\AtBeginDocument{%
  }

\setcopyright{acmlicensed}
\copyrightyear{2026}
\acmYear{2026}
\acmDOI{XXXXXXX.XXXXXXX}
\acmConference[WWW '26]{The Web Conference 2026}{April 13--17, 2026}{Dubai, United Arab Emirates}
\acmISBN{978-1-4503-XXXX-X/2018/06}

\usepackage{amsmath}
\usepackage{graphicx}
\usepackage{subcaption}
\usepackage{multirow}
\usepackage{enumitem}

\begin{document}

\newcommand{\tighttt}[2][8pt]{
  {\fontsize{#1}{#1}\selectfont
   \ttfamily
   \spaceskip=0.3em\relax
   \kern-0.2em#2\kern-0.2em
  }
}

\title{When Rules Fall Short: Agent-Driven Discovery of Emerging Content Issues in Short Video Platforms}

\author{Chenghui Yu}
\authornote{Equal contribution.}
\affiliation{
  \institution{TikTok Inc.}
  \city{San Jose, CA}
  \country{USA}}
\email{yuchenghui@tiktok.com}

\author{Hongwei Wang}
\authornotemark[1]
\affiliation{
  \institution{TikTok Inc.}
  \city{San Jose, CA}
  \country{USA}}
\email{hongwei.w@tiktok.com}

\author{Junwen Chen}
\affiliation{
  \institution{TikTok Inc.}
  \city{San Jose, CA}
  \country{USA}}
\email{chenjunwen.javen@tiktok.com}

\author{Zixuan Wang}
\affiliation{
  \institution{TikTok Inc.}
  \city{San Jose, CA}
  \country{USA}}
\email{zixuan.wang1@tiktok.com}

\author{Bingfeng Deng}
\affiliation{
  \institution{TikTok Inc.}
  \country{Singapore}}
\email{dengbingfeng@tiktok.com}

\author{Zhuolin Hao}
\affiliation{
  \institution{TikTok Inc.}
  \city{San Jose, CA}
  \country{USA}}
\email{haozhuolin@tiktok.com}

\author{Hongyu Xiong}
\affiliation{
  \institution{TikTok Inc.}
  \city{San Jose, CA}
  \country{USA}}
\email{hongyu.xiong@tiktok.com}

\author{Yang Song}
\affiliation{
  \institution{TikTok Inc.}
  \city{San Jose, CA}
  \country{USA}}
\email{ys@sonyis.me}

\renewcommand{\shortauthors}{C. Yu, H. Wang et al.}

\begin{abstract}
    Trends on short-video platforms evolve at a rapid pace, with new content issues emerging every day that fall outside the coverage of existing annotation policies.
    However, traditional human-driven discovery of emerging issues is too slow, which leads to delayed updates of annotation policies and poses a major challenge for effective content governance.
    In this work, we propose an automatic issue discovery method based on multimodal LLM agents.
    Our approach automatically recalls short videos containing potential new issues and applies a two-stage clustering strategy to group them, with each cluster corresponding to a newly discovered issue.
    The agent then generates updated annotation policies from these clusters, thereby extending coverage to these emerging issues.
    Our agent has been deployed in the real system.
    Both offline and online experiments demonstrate that this agent-based method significantly improves the effectiveness of emerging-issue discovery (with an F1 score improvement of over 20\%) and enhances the performance of subsequent issue governance (reducing the view count of problematic videos by approximately 15\%).
    More importantly, compared to manual issue discovery, it greatly reduces time costs and substantially accelerates the iteration of annotation policies.
\end{abstract}

%% The code below is generated by the tool at http://dl.acm.org/ccs.cfm.



\maketitle

    \begin{figure}[t]
        \setlength{\belowcaptionskip}{-12pt}
        \begin{subfigure}{0.45\textwidth}
            \centering
            \includegraphics[width=\linewidth]{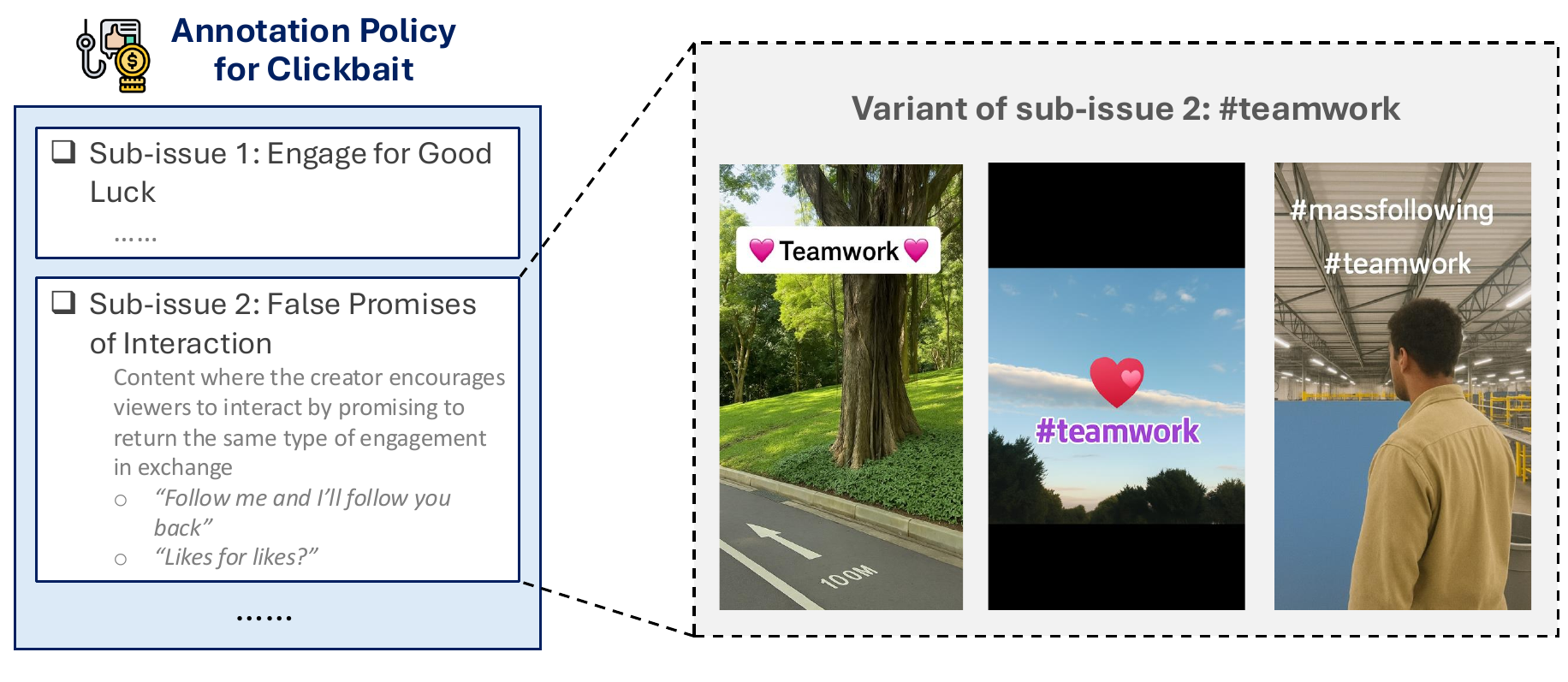}
            \caption{A variant of an existing sub-issue in \textsf{Clickbait}}
            \label{fig:example_1}
        \end{subfigure}
        \vfill
        \begin{subfigure}{0.45\textwidth}
            \centering
            \includegraphics[width=\linewidth]{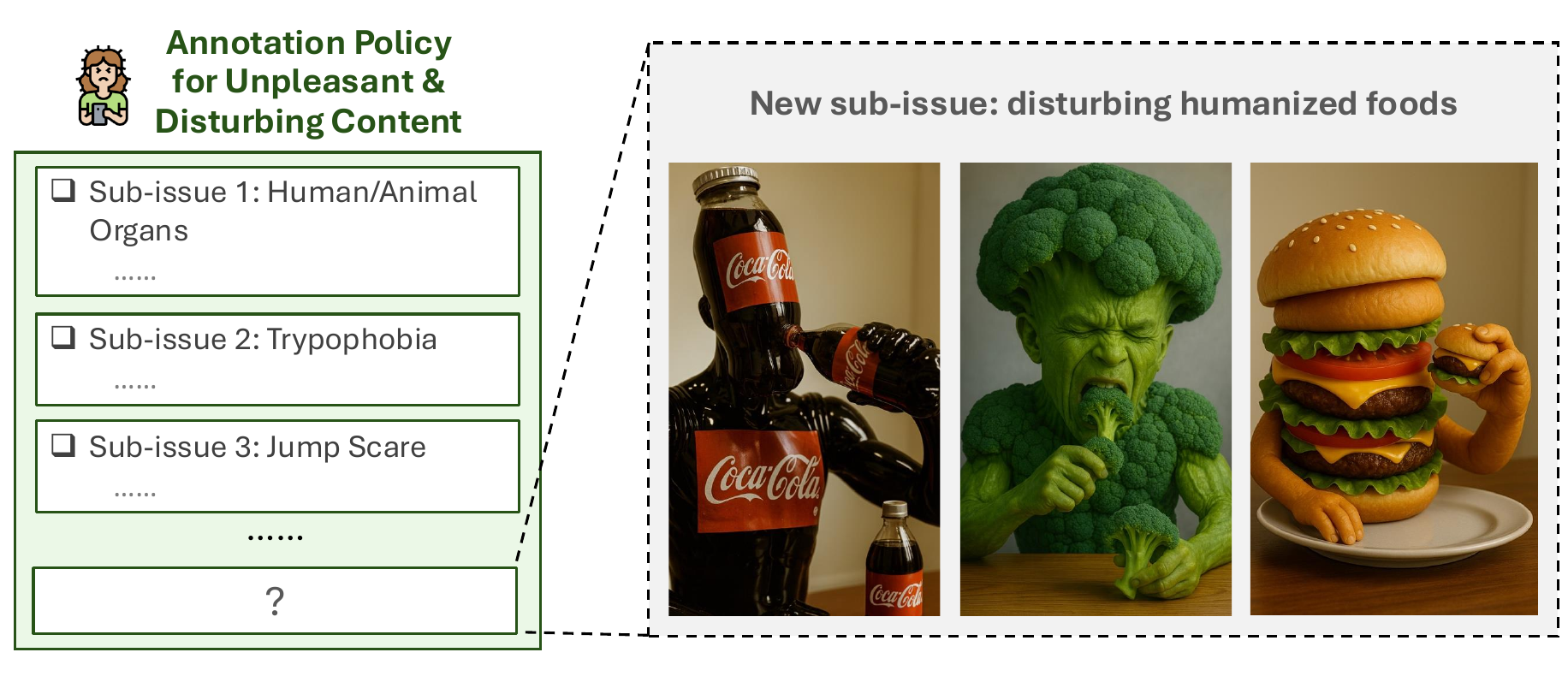}
            \caption{A new sub-issue in \textsf{Unpleasant \& Disturbing Content}}
            \label{fig:example_2}
        \end{subfigure}
        \caption{Examples of emerging content issues in short videos uploaded by users. See Section \ref{sec:introduction} for more details.}
        \label{fig:example}
    \end{figure}

    \begin{figure*}[t]
        \setlength{\belowcaptionskip}{-4pt}
        \centering
        \includegraphics[width=\textwidth]{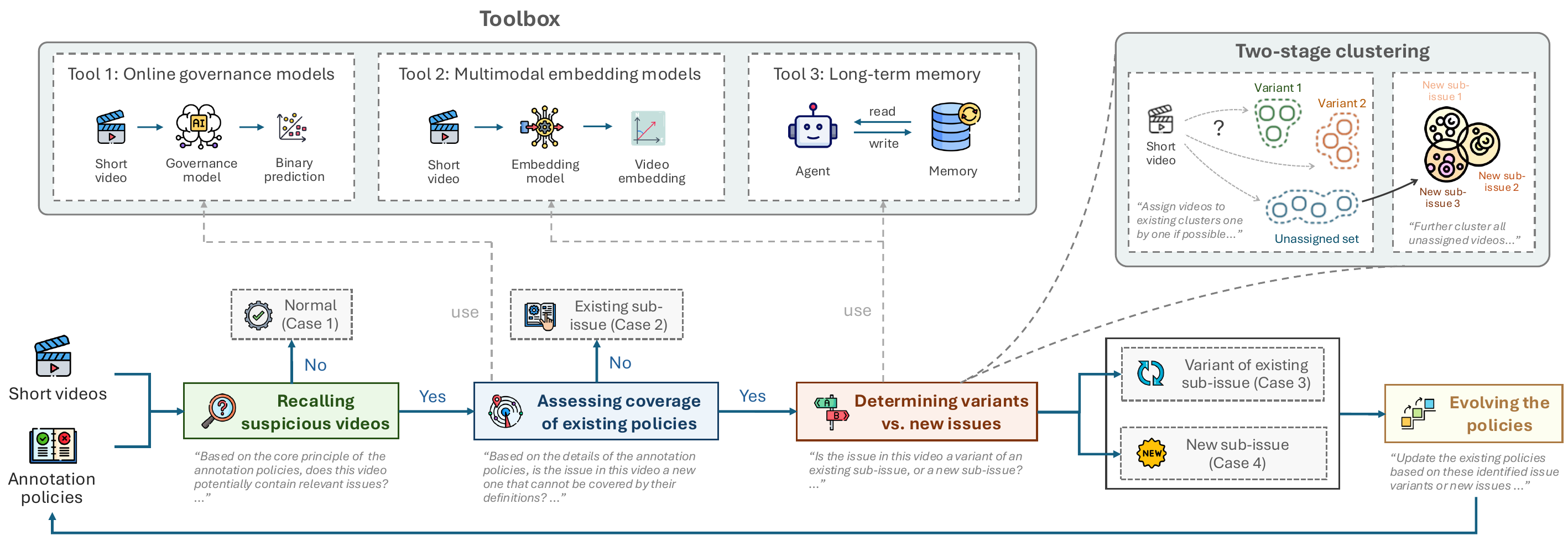}
        \caption{Illustration of the issue discovery agent. Given a short video and the annotation policies for a specific issue, the agent follows a four-phase process to categorize videos into four cases (Cases 1 – 4), and finally produces an updated version of the annotation policies. The agent can autonomously choose to invoke certain tools. See Section \ref{sec:approach} for more details.}
        \label{fig:framework}
    \end{figure*}

\section{Introduction}
    \label{sec:introduction}
    Short videos uploaded by users on TikTok may contain various \textit{content issues}, such as sexually suggestive content, non-original content, clickbait, or unpleasant \& disturbing content.
    These problematic videos may negatively impact the users' FYF(For You feed) experience and the sustainable development of the community\cite{tiktok_guidelines_2025}.
    To address this, our platform employs \textit{governance models} to detect whether videos include such issues.
    Typically, the platform first establishes detailed \textit{annotation policies} for identifying each specific issue.
    These annotation policies are then used to train human annotators who label a set of videos accordingly.
    Finally, the labeled dataset is used to train governance models, which can automatically identify problematic videos at scale.

    From the perspective of content governance, the relationship between the platform and users resembles a ``cat-and-mouse'' dynamic.
    The platform strives to design broader and more detailed policies to cover more issues, while users, whether intentionally or unintentionally, upload videos that contain variants of existing issues or entirely new ones, thereby evading the scope of the policies.
    Figure \ref{fig:example} illustrates two examples:
    (1) As shown in Figure \ref{fig:example_1}, within the \textsf{Clickbait} issue and specifically the \textsf{False Promises of Interaction} sub-issue, we observe a new emerging variant: using stickers with ``\textsf{\#teamwork}'' to induce likes or follows.
    However, the keyword ``\textsf{teamwork}'' is not included in the definition of this sub-issue.
    (2) As shown in Figure \ref{fig:example_2}, under the issue of \textsf{Unpleasant \& Disturbing Content}, we identify a new sub-issue: AI-generated content that morphs food into human-like forms, often paired with eating sound effects and scenes of consuming the same items.
    Such visuals can be disturbing, but they do not fit into the definition of any existing sub-issue.

    Trends on TikTok evolve at an extremely rapid pace.
    A new video may appear today, and within just a few days, the platform can be flooded with countless imitations.
    Therefore, promptly identifying emerging content issues and establishing new annotation policies is a critical task for the platform.
    At present, the process is entirely \textit{manual}:
    First, based on user reports and proactive screening by moderators, a set of candidate videos suspected of containing emerging issues is collected.
    Next, the content governance team assesses whether these videos have sufficient impact and potential harm.
    If so, new annotation policies are created for the new issue, or existing policies are updated to cover the identified variant.
    However, this manual workflow is highly time-consuming, often taking several months to complete, which leaves the platform reacting very slowly to new issues.
    Moreover, relying solely on human judgment introduces strong subjectivity with inconsistent criteria, resulting in suboptimal annotation policies.

    To overcome the limitations of the fully manual process, one promising approach is to introduce machine learning models for automatic discovery of new issues.
    However, a clear challenge arises: since the target is previously undefined issues, we typically have few or no labeled samples available, making supervised learning methods inapplicable.
    This means the models must rely largely on analyzing the content of the videos themselves and understanding annotation policies to make decisions.
    These constraints lead us to explore \textit{MLLM-based agents} as a potential solution.
    MLLMs bring extensive world knowledge, and the agent framework allows them to leverage planning and reasoning capabilities for complex tasks.
    This paper introduces our progress in \textit{applying the agent approach to emerging issue discovery}.

    As shown in Figure \ref{fig:framework}, the workflow of our agent can be divided into four phases:
    \begin{itemize}
        \item
            \textit{Recalling suspicious videos}.
            Given a short video and the annotation policies for a specific issue, the agent first determines whether the video is likely to contain that issue by any chance.
            If the answer is no, the video is classified as normal and filtered out.
        \item
            \textit{Assessing the coverage of existing policies}.
            The agent then checks whether the issue present in the video is already covered by the existing annotation policies.
            If it is, the video is categorized as an instance of an existing sub-issue and filtered out.
            At this stage, the agent may also use \textit{governance models} as auxiliary tools to assist in the judgment.
        \item
            \textit{Determining variants vs. new issues}.
            Here, the agent decides whether the video represents a variant of an existing sub-issue or introduces an entirely new sub-issue.
            To achieve this, the agent applies a \textit{two-stage clustering} method to a batch of input videos, which assigns each video either to an existing cluster or to a newly created cluster.
            In addition, the agent may employ two tools: an \textit{embedding} tool and a \textit{memory} tool, to support its decision.
        \item
            \textit{Evolving the policies}.
            Once the clusters for variants and new sub-issues are formed, the agent summarizes them and generates updated policies.
            These updated policies are used to guide human annotators to label new data for training governance models.
            They can also be iteratively refined by the issue discovery agent in real-world deployment environments.
    \end{itemize}

    We conduct offline experiments on two content issues.
    The results demonstrate that, compared with supervised fine-tuning based approaches, our agent-based method achieves significant improvements in the issue discovery task.
    Specifically, the best-performing agent achieves absolute improvements in macro-F1 of 26.3\% and 22.1\%, respectively, on the two content issues, compared with the best-performing SFT model.
    Online A/B testing further shows that the annotation policies generated by the agent achieve substantial gains in issue governance, reducing 14.9\% views of videos containing new issues.
    More importantly, compared with manual issue discovery, our approach dramatically reduces time costs, making the entire process more efficient.

\section{Related Work}
    \noindent\textbf{Video Content Governance}.\ \
        With the exponential growth of user-generated video content, video content governance has become critical for ensuring safe digital ecosystems and meeting regulatory requirements.
        Early \textit{rule-based} solutions rely on predefined criteria such as keyword matching, OCR for on-screen text, and heuristic visual filters \cite{youtube2023guidelines,bhagoji2024community}.
        Subsequently, \textit{machine learning based} models are introduced \cite{li2020multiscale,dosovitskiy2021image}.
        \textit{Hybrid} approaches have also emerged, which integrate rule-based checks with learning models to balance efficiency and nuance \cite{facebook2022aisafety}.
        Recent work also focuses on the model efficiency \cite{hosseini2023faster}, multimodal refinement \cite{ahmed2024enhanced}, LLM-driven governance \cite{aldahoul2024advancing, wang-etal-2025-filter}, and industrial solutions \cite{yu2024usm, yu2025unified,oak2025reranking}.
    
    \vspace{0.5em}
    \noindent\textbf{Knowledge Discovery}.\ \
        Knowledge discovery is the process of extracting actionable insights, patterns, and structured knowledge from unstructured or semi-structured data \cite{jiang2023finance,choi2024healthcare}.
        Early approaches \cite{agrawal1994fast, han2023traditional} relied on traditional data mining techniques, which are effective for structured data but struggled to capture the semantic nuances of unstructured or multi-modal data.
        More recent work has focused on multi-modal information integration \cite{wang2023multi} and its combination with LLMs \cite{rajagopal2023llm, zhang2024llm}.
        Despite these advances, several critical challenges remain, for example, modality bias \cite{wang2023multi} and the need for validation to address LLM hallucinations \cite{zhang2024llm}.
        
    \vspace{0.5em}
    \noindent\textbf{LLM-Based Agents}.\ \
        The rise of LLMs has fueled the development of LLM-based agents, which surpass traditional rule-driven or command-based systems (e.g., early assistants like Siri) by leveraging LLMs' advanced language understanding and generation capabilities \cite{wooldridge2022foundations}.
        Recent research on LLM-based agents centers on three directions:
        (1) \textit{Reasoning enhancement}.
        Techniques such as Chain-of-Thought \cite{wei2022chain} or Tree-of-Thoughts \cite{yao2023tree} prompts guide LLMs to break down complex problems into sequential or multi-sequential steps.
        (2) \textit{Tool integration}.
        Frameworks like ReAct \cite{yao2022react}, Toolformer \cite{schuurmans2023toolformer}, WebGPT \cite{nakano2021webgpt}, and HuggingGPT \cite{shen2023hugginggpt} demonstrate how agents can interact with external tools (APIs, databases) through action generation, bridging abstract reasoning and real-world execution.
        (3) \textit{Optimization frameworks}.
        Parameter-driven methods (e.g., supervised fine-tuning, reinforcement learning) refine model weights to better align agent behavior with task goals, while parameter-independent approaches (e.g., external knowledge retrieval, tools calling) improve performance without retraining, offering efficiency benefits for large-scale deployment \cite{zhang2024optimization}.

\section{Problem Formulation}
    \label{sec:problem_formulation}
    Suppose we are interested in a particular content issue $I$ (multi-issue scenarios will be discussed in Section \ref{sec:deployment}).
    Issue $I$ has an associated annotation policy $P_I$, which provides a detailed definition of $I$.
    $I$ consists of $n$ sub-issues: $I = \{ 1, 2, \cdots, n\}$.
    Correspondingly, the policy $P_I$ is also divided into $P_I = \{ P_1, P_2, \cdots, P_n\}$, where each $P_i$ contains the textual definition of sub-issue $i$. 

    Given a batch of short videos $\mathcal V$, we want to determine for each video $v \in V$ which of the following cases it falls into:
    \begin{itemize}
        \item
            \textit{Negative} (Case 1):
            $v$ is normal and does not contain any content related to issue $I$.
        \item
            \textit{Positive, existing sub-issue} (Case 2):
            $v$ belongs to one of the existing sub-issue $i \in I$.
            In this case, the specific violation of $v$ is explicitly defined in $P_i$.
        \item
            \textit{Positive, variant of existing sub-issue} (Case 3):
            $v$ still belongs to a sub-issue $i \in I$, but its specific violation is not explicitly covered in $P_i$. Instead, it represents a new variant.
        \item
            \textit{Positive, new sub-issue} (Case 4):
            $v$ does not belong to any existing sub-issue of $I$ and represents a new sub-issue.
            For the entire batch $\mathcal V$, the number of newly discovered sub-issues may range from 0 (no video involves a new sub-issue) to $|\mathcal V|$ (every video corresponds to a distinct new sub-issue).
    \end{itemize}

    For the videos identified as Case 3 or 4, we further need to perform sub-issue-level identification.
    This allows us to summarize their underlying patterns and subsequently refine or update the existing annotation policy.

\section{Approaches}
    \label{sec:approach}
    \subsection{Recalling Suspicious Videos}
        Given a short video $v$, the agent's first step is to determine whether the content of $v$ contains issue $I$.
        In other words, this phase is essentially a binary classification task: Case 1 vs. Cases 2/3/4.
        Although the annotation policy $P_I$ provides a detailed definition of issue $I$ that the agent can leverage, this phase is still nontrivial.
        The reason is that its purpose is to recall videos for issue discovery, which means that the agent must not only recall videos already covered by $P_I$, but also those that contain variants of existing sub-issues or new sub-issues of $I$.

        To achieve this, the key strategy is not to rely heavily on the literal policy descriptions, but instead to capture the \textit{essential decision logic} of the issue.
        For example, in \textsf{Clickbait}, the essential logic consists of three principles:
        (1) \textit{Engagement authenticity}: engagement must come from genuine user interest, not manipulation or deception.
        (2) \textit{Creator fairness}: artificially inflated engagement undermines fairness for other creators.
        (3) \textit{User experience}: misleading tactics make the content diverge from user expectations, eroding trust and harming user experience.
        Thus, when making judgments, the agent must evaluate along these dimensions:
        ``\tighttt{Does user engagement with the video reflect genuine interest?}
        \tighttt{Does the video unfairly disadvantage other creators?}
        \tighttt{Does the video degrade user experi- ence?}''
        If none of these essential logics apply, the video is classified as normal.
        Otherwise, it is considered problematic and proceeds to the next phase in the agent's workflow.

    \subsection{Assessing Coverage of Existing Policies}
        Through the previous phase, the agent filtered out normal videos (Case 1) and retained those containing issue $I$.
        The task in this phase is to further exclude the videos that are already covered by the annotation policy $P_I$, leaving only those that involve variants of existing sub-issues or new sub-issues.
        In other words, this is a binary classification task distinguishing Case 2 from Cases 3/4.
        
         To achieve this, the agent carefully consults the textual definition $P_i$ for each sub-issue $i \in I$, in order to make precise judgments.
        A typical prompt may look like: ``\tighttt{Based on the following annotation policy, determine whether the input video falls into any of the \\listed cases. Sub-issue 1: <definitions>. Sub-issue 2: <defini-\\tions>... Input video: <frames> <additional\_text\_info>}.''

        \subsubsection{Tool 1: Governance Models}
            In this phase, the agent can also leverage governance models as auxiliary tools to support its judgment.
            Governance models are lightweight and specialized classifiers deployed online for detecting specific issues:
            \begin{equation}
                \mathcal G_I(v; P_I): \mathcal V \rightarrow \{0, 1\},
            \end{equation}
            where $\mathcal G_I$ denotes the governance model for issue $I$.
            These models are trained on human-labeled data under the guidance of existing policies $P_I$.
            As a result, governance models are capable of determining whether the video falls under $P_I$, thereby demonstrating the ability to distinguish Case 2 from Cases 1/3/4.
            Since Case 1 videos have already been filtered out in Phase 1, governance models in this phase can effectively differentiate Case 2 from Cases 3/4.
            Specifically, when the agent has low confidence in its own decision, it can autonomously invoke a governance model to estimate the probability that the given video should be classified as normal, and use this signal as a reference to refine its judgment.

    \subsection{Determining Variants vs. New Issues}
        For problematic videos that are not covered by the existing policies, we need to further determine whether they are variants of existing sub-issues or new sub-issues, that is, distinguishing between Case 3 and Case 4.
        This is challenging because (1) the number and definitions of potential new sub-issues are unknown, and (2) video batches are usually large in practice, making it difficult for an agent to process all of them at once.
        
        To address this, we design a \textit{two-stage clustering} approach \cite{zubarouglu2021data,o2002streaming}.
        Given a batch of videos $\mathcal V$ that has passed the previous two filtering phases, we cluster the videos at the sub-issue level through a \textit{streaming clustering} stage and an \textit{offline clustering} stage.
        The procedure of streaming clustering is as follows:
        \begin{itemize}
            \item
                \textit{Initialization}.
                Based on the sub-issue definitions in the existing policies $P_I$, we initialize the clusters of known sub-issues:
                \begin{equation}
                    \{ c_1, c_2, \cdots, c_n\} = INIT(P_I),
                \end{equation}
                where each $c_i$ is the synopsis (compressed summary) of sub-issue $i$ and $n$ is the number of existing sub-issues.
        \end{itemize}
        
        For each video $v \in \mathcal V$, we then perform the following steps:
        \begin{itemize}
            \item
                \textit{Assignment}.
                We select the cluster $c_i$ to which $v$ most likely belongs, and add $v$ to $c_i$.
                $v$ can therefore be seen as an instance of sub-issue $i$'s variant.
                If it is determined that $v$ does not belong to any cluster, we put $v$ into a temporary cluster $c_*$ consisting of all unassigned videos.
                This step can be expressed as
                \begin{equation}
                    c_i = ASSIGN \big( v; \{ c_1, c_2, \cdots, c_n, c_* \} \big).
                \end{equation}
            \item
                \textit{Updating}.
                After video $v$ is assigned to cluster $c_i$, $c_i$ should update its synopsis to incorporate the new information if $c_i \neq c_*$:
                \begin{equation}
                    c_i = UPDATE(c_i; v).
                \end{equation}
        \end{itemize}

        After completing the streaming clustering stage, the videos in $c_*$ can be considered as not belonging to existing sub-issue variants but representing new sub-issues.
        For all videos in $c_*$, we then perform another round of offline clustering to group them:
        \begin{equation}
            \{ c_{n+1}, \cdots, c_{n+m} \} = OFFLINE\_CLUSTERING(c_*),
        \end{equation}
        where $m$ is a hyperparameter denoting the number of newly discovered sub-issues.

        Building on this framework, we design two specific tool implementations for agents to autonomously choose to use: an \textit{embedding} tool and a \textit{memory} tool.

        \begin{figure*}[t]
            \setlength{\belowcaptionskip}{-3pt}
            \centering
            \includegraphics[width=.98\linewidth]{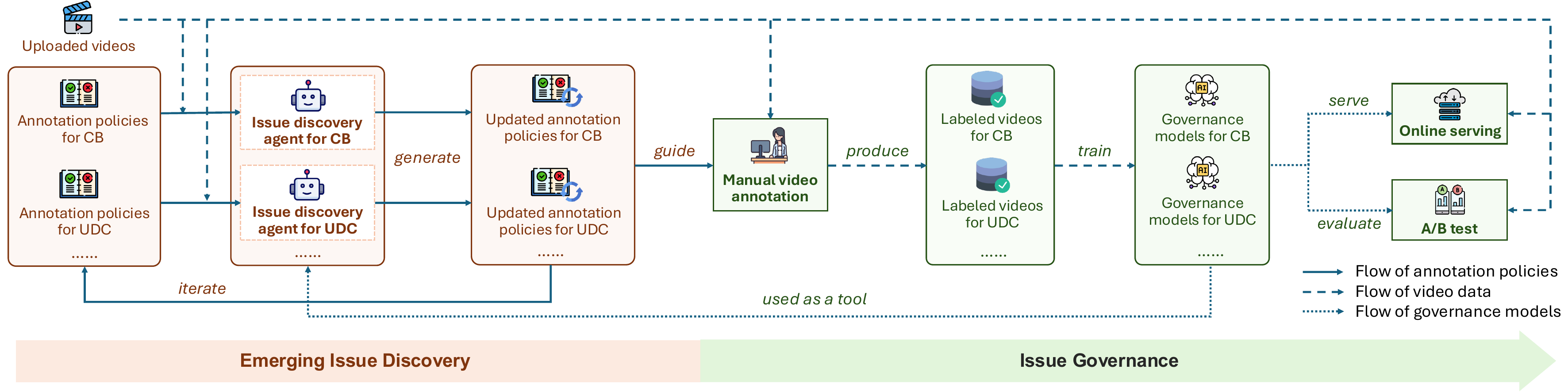}
            \caption{The flowchart of real-world deployment for emerging issue discovery and issue governance. We use Clickbait (CB) issue and Unpleasant \& Disturbing Content (UDC) issue as examples. See Section \ref{sec:deployment} for more details.}
            \label{fig:deployment}
        \end{figure*}
        
        \subsubsection{Tool 2: Multimodal Embedding}
        This tool uses a multimodal embedding model (e.g., CLIP \cite{radford2021learning}) to compute representation vectors for both short videos and policy texts.
        The similarity between embeddings is then used as the criterion for clustering.
        
        Specifically, during the initialization step, for each sub-issue $i \in I$, we use the multimodal embedding model $\mathcal M$ to compute the embedding of its textual definition $P_i$, which is used as the synopsis of the cluster for sub-issue $i$:
        \begin{equation}
            c_i = \mathcal M (P_i), \quad for \ i \in I.
         \end{equation}
        
        In the assignment step, we first compute the cosine similarity between the current video $v$ and the synopsis of all clusters $\{ c_1, \cdots, c_n \}$, and then take the maximum similarity:
        \begin{equation}
        \label{eq:embedding_select}
            max\_sim = \max\nolimits_{k=1}^n \textrm{cos\_sim} \big( \mathcal M (v), c_k \big),
        \end{equation}
        where $\mathcal M(v)$ computes the embedding of $v$ by averaging the embedding of its video frames and the embedding of its additional textual information (such as OCR and ASR results).
        If $max\_sim \geq \delta$ where $\delta \in [-1, 1]$ is a predefined threshold, $v$ is assigned to the corresponding cluster.
        Otherwise, we consider $v$ dissimilar to all clusters and assign $v$ to $c_*$.

        In the updating step, the cluster $c$ assigned with $v$ needs to recompute its centroid if $c \neq c_*$:
        \begin{equation}
            c = \frac{1}{C + 2} \mathcal M (v) + \frac{C + 1}{C + 2} c,
        \end{equation}
        where $C$ is the number of videos already assigned to $c$.

        Finally, in the offline clustering stage, we use off-the-shelf algorithms such as K-means \cite{ahmed2020k} or hierarchical clustering \cite{murtagh2012algorithms} to group videos in $c_*$ into new sub-issues.

        \subsubsection{Tool 3: Long-Term Memory}
        The memory tool provides the agent with a long-term memory \cite{wulongmemeval,wang2023augmenting} module, which is designed to store clustering information generated while processing batches of videos.

        During the initialization step, the agent analyzes the annotation policy $P_i$ for each sub-issue $i \in I$ and produces a summary.
        This summary becomes the synopsis $c_i$ for the corresponding cluster, which is then written into memory:
        \begin{equation}
        \begin{split}
            &c_i = agent.summarize(P_i),\\
            &agent.write\_memory(c_i), \quad for \ i \in I.
        \end{split}
        \end{equation}

        In the assignment step, for a given video $v$, the agent retrieves all stored cluster synopses from memory and selects the one most closely matching $v$:
        \begin{equation}
        \label{eq:agent_select}
        \begin{split}
            &\{ c_1, \cdots, c_n \} = agent.read\_memory(),\\
            &c_i = agent.select \big( \{ c_1, \cdots, c_n \}; v \big).
        \end{split}
        \end{equation}
        If no suitable cluster is found, the agent assigns $v$ to $c_*$.

        In the update step, the agent re-summarizes the selected cluster's synopsis based on the content of $v$, then writes the updated version back to memory:
        \begin{equation}
            c_i = agent.summarize \big( \{c_i, v\} \big), \quad agent.write\_memory(c_i).
        \end{equation}

        Finally, in the offline clustering stage, the agent performs a one-time clustering of all videos in $c_*$, without relying on embedding-based methods such as K-means.
        This is feasible because the size of $c_*$ is already much smaller than that of $\mathcal V$:
        \begin{equation}
            \{ c_{n+1}, \cdots, c_{n+m} \} = agent.cluster(c_*).
        \end{equation}

    \subsection{Evolving the Policies}
        After completing sub-issue-level clustering for all videos, the final phase is to generate new annotation policies.
        
        For clusters $c_1, \cdots, c_n$ that correspond to existing sub-issues, the newly added videos within these clusters can be regarded as variants of those sub-issues.
        The agent summarizes these videos and updates the existing policy accordingly based on the old policy $P_i$.
        This includes producing a refined definition and selecting the most representative examples from the new videos:
        \begin{equation}
            P_i = agent.update\_policy(P_i, c_i), \quad if \ i \leq n.
        \end{equation}

        For newly formed clusters $c_{n+1}, \cdots, c_{n+m}$, these can be treated as newly discovered sub-issues. Similarly, the agent summarizes the videos in each cluster and generates a new policy definition and representative examples:
        \begin{equation}
            P_i = agent.create\_policy(c_i), \quad if \ i > n.
        \end{equation}

        Note that Phase 1 and Phase 2 of the agent's pipeline may mistakenly recall some Case 1 or Case 2 videos.
        %As a result, the final clustering output may include clusters of normal videos or clusters already covered by existing policies.
        Therefore, any policies generated by the agent must undergo necessary human review.
        Nevertheless, compared with manual issue discovery, this agent-based approach significantly reduces the time cost and accelerates the iteration of subsequent issue governance.
        We will further discuss efficiency improvements in Section \ref{sec:deployment}.

    \begin{table*}[t]
        \centering
        \setlength{\tabcolsep}{8pt}
        \small
        \begin{tabular}{c|cccc|cccc}
            \toprule
            \multirow{2}{*}{\textbf{Methods}} & \multicolumn{4}{c|}{\textbf{CB}} & \multicolumn{4}{c}{\textbf{UDC}} \\
            & \textit{Precision} & \textit{Recall} & \textit{Weighted-F1} & \textit{Macro-F1} & \textit{Precision} & \textit{Recall}& \textit{Weighted-F1} & \textit{Macro-F1} \\
            \midrule
            \midrule
            \multicolumn{1}{c}{\textbf{Issue discovery agent}} & \multicolumn{8}{c}{} \\
            \makebox[14em][l]{\textit{No tools}} & 39.85 & 47.70 & 88.96 & 36.83 & 50.41 & 45.90 & 89.77 & 45.75 \\
            \makebox[14em][l]{\quad+ \textit{Tool 1 (governance models)}} & 51.95 & 59.98 & 92.33 & 54.64 & 51.02 & 49.17 & 90.31 & 47.09 \\
            \makebox[14em][l]{\quad+ \textit{Tool 2 (embedding models)}} & 46.52 & 47.95 & 89.06 & 37.38 & 49.65 & \underline{52.25} & 89.27 & \underline{48.56}  \\
            \makebox[14em][l]{\quad+ \textit{Tool 3 (long-term memory)}} & 41.52 & 47.95 & 89.04 & 37.26 & 49.46 & 51.85 & 89.23 & 48.30  \\
            \makebox[14em][l]{\quad+ \textit{Tools 1 \& 2}} & \textbf{59.08} & \textbf{70.49} & \textbf{92.85} & \textbf{62.91} & \textbf{67.40} & \textbf{57.92} & \textbf{90.89} & \textbf{58.83} \\
            \makebox[14em][l]{\quad+ \textit{Tools 1 \& 3}} & \underline{55.20} & \underline{62.03} & \underline{92.58} & \underline{57.80} & \underline{52.00} & 46.75 & 89.77 & 47.36 \\
            \midrule
            \multicolumn{1}{c}{\textbf{Supervised fine-tuning}} & \multicolumn{8}{c}{} \\
            %\multicolumn{1}{l}{\textit{8 training samples for Cases 3/4}} &  &  &  & \multicolumn{1}{c}{} &  &  &  &  \\
            \makebox[14em][l]{\textit{w/ REDA as base model}} & 32.99 & 45.15 & 89.83 & 36.88 & 45.87 & 33.71 & \underline{90.73} & 36.71 \\
            \makebox[14em][l]{\textit{w/ Qwen2.5-VL as base model}} & 32.95 & 45.14 & 89.80 & 36.26 & 40.23 & 32.13 & 90.24 & 34.19 \\
            %\midrule
            %\midrule
            %\multicolumn{1}{l}{\textit{Full training samples for Cases 3/4}} &  &  &  & \multicolumn{1}{c}{} &  &  &  &  \\
            %\makebox[14em][l]{\quad \textit{w/ REDA}} & 37.80 & 49.41 & 93.05 & 41.55 & 68.09 & 57.14 & 95.36 & 59.36 \\
            %\makebox[14em][l]{\quad \textit{w/ Qwen2.5-VL}} & 36.97 & 49.09  & 92.93 & 40.80 & 67.23 & 55.90 & 95.04 & 56.49 \\
            \bottomrule
        \end{tabular}
        \vspace{1ex}
        \caption{Result of agent and SFT on the four-class classification task. The evaluation metrics are Precision, Recall, Weighted-F1, and Macro-F1 score (in \%). The highest value in each column is shown in bold, and the second-highest is underlined.}
        \vspace{-4ex}
        \label{tab:main}
    \end{table*}
    
\section{Deployment}
    \label{sec:deployment}
    Our issue discovery agent has already been deployed in the product environment.
    Figure \ref{fig:deployment} illustrates the flowchart of emerging issue discovery and issue governance, as well as how the modules interact.
    Three points are worth emphasizing:
    \begin{itemize}
        \item
            \textit{Multiple issues}.
            In Section \ref{sec:problem_formulation}, we simplify the problem to discovery within a single issue category.
            In practice, the platform needs to monitor dozens of issue categories, and we don't know in advance which issues a given video might contain.
            Therefore, we need to design a dedicated agent for each type of issue and perform parallel processing on each video for better efficiency.
        \item
            \textit{Ultimate evaluation metrics for issue discovery}.
            Evaluating the effectiveness of issue discovery is not a straightforward task.
            A strong issue discovery agent should be able to identify more potential issues → generate more comprehensive annotation policies → guide the annotation of higher-quality data → ultimately enable the training of stronger governance models.
            For this reason, we consider the performance of governance models to be the ultimate metric for assessing an issue discovery agent.
            Accordingly, we conduct A/B tests on governance models at the final stage of content governance.
            Detailed experimental results are presented in Section \ref{sec:ab_test}.
        \item
            \textit{Efficiency improvements}.
            The issue discovery agents replace the previous manual process of issue discovery.
            This significantly reduces labor and time costs, accelerates policy iteration, and also greatly speeds up subsequent processes of manual video annotation and governance model training.
            We will provide more results and analysis of these efficiency improvements in Section \ref{sec:efficiency}.
    \end{itemize}

\section{Experiments}
    \label{sec:experiments}
    \subsection{Experimental Setup}
        \subsubsection{Content Issues}
        We conducted experiments on two issues:
        \begin{itemize}
            \item
                \textit{Clickbait (CB)}, which refers to manipulative practices that prompt viewers to interact with content in ways that artificially inflate engagement metrics and compromise the integrity of the recommendation system, leading to unfair exposure of content that would not otherwise be merited.
                The existing annotation policy for CB contains 12 sub-issues.
            \item
                \textit{Unpleasant \& Disturbing Content (UDC)}, which refers to content that may evoke uncomfortable feelings (e.g., fear, disgust, or shock) through imagery, audio, or editing, thereby disrupting the user's viewing experience.
                The existing annotation policy for UDC contains 19 sub-issues.
        \end{itemize}

        \subsubsection{Datasets}
        Each video contains both textual and visual information.
        Textual information includes the title, stickers, OCR results, and ASR transcripts.
        Visual information consists of 8 uniformly sampled video frames, each with a resolution of 336 $\times$ 336.
        
        For the CB issue, the evaluation set contains 24,670 samples, of which 5.3\% belong to Cases 3/4.
        For the UDC issue, the evaluation set contains 21,890 samples, of which 4.1\% belong to Cases 3/4.
        Samples in Cases 3/4 include 6 sub-issues.

        \subsubsection{Hyper-Parameter Settings}
        We used our pretrained model REDA \cite{wang2025reasoning} as the base model for the agent, with a size of 7B.
        For Tool 1, governance models are SFT models built on top of VLMo \cite{bao2022vlmo} for CB and LLaVA \cite{liu2023visual} for UDC.
        For Tool 2, the embedding model is based on OpenCLIP ViT-L-14 \cite{cherti2023reproducible}; the assignment threshold $\delta$ is set to 0.4; the number of clusters in offline clustering is set to 4.
        These optimal values are determined through a search process.

        \subsubsection{Baseline Methods}
        We perform supervised fine-tuning (SFT) on the MLLM and take this as our baseline method.
        The base models for SFT are our pretrained model \textit{REDA} 7B \cite{wang2025reasoning} and native \textit{Qwen2.5-VL} 7B \cite{bai2025qwen2}.
        We apply LoRA-based \cite{hu2022lora} fine-tuning by adding rank-168 trainable matrices into the transformation matrices of the feed-forward modules at every layer of both the visual encoder and the LLM.
        In addition, we add a classification head on top of the final layer, implemented as an MLP with one hidden layer.
        In total, the number of trainable parameters is 441M.

        The size of the SFT training sets is as follows.
        CB issue: 10k Case-1 samples, 4k Case-2 samples, and between 8 to 512 samples for Cases 3/4.
        UDC issue: 10k Case-1 samples, 2k Case-2 samples, and between 8 to 150 samples for Cases 3/4.
        The default number of Case-3/4 samples is set to 64, which simulates the real scenario where we have only a very small amount of human-annotated videos for new issues.
        In this setting, SFT can be fairly compared against agent (but note that SFT still has an advantage as the agent does not use any training data).
        When the number of Case-3/4 samples exceeds 64, the goal is instead to explore the upper bound of SFT performance.

    \begin{figure*}[t]
		\centering
		\begin{minipage}[t]{0.31\textwidth}
            \includegraphics[width=\textwidth]{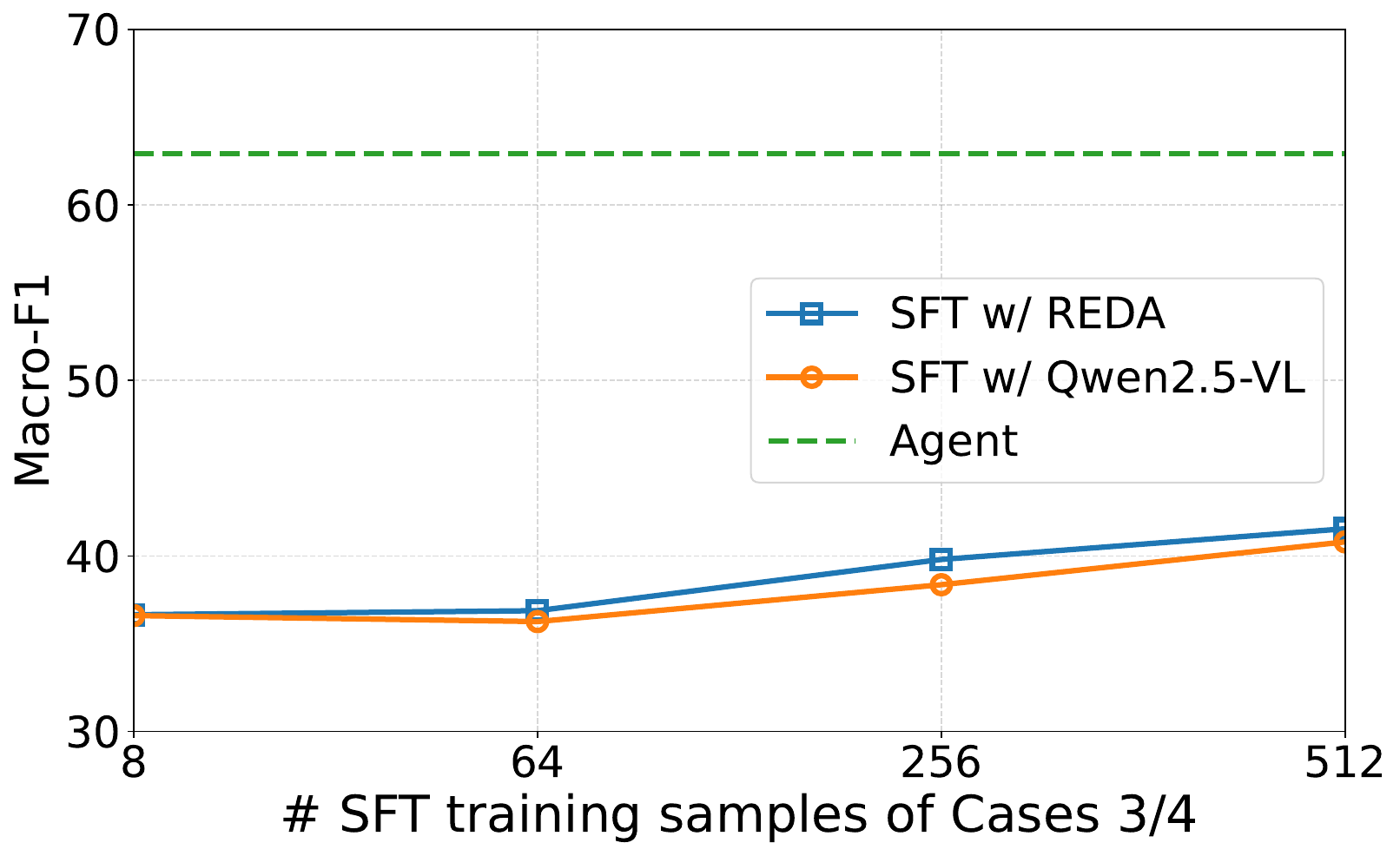}
            \caption{Impact of \# SFT training samples of Cases 3/4 on SFT performance.}
            \label{fig:sft}
        \end{minipage}
        \hfill
        \begin{minipage}[t]{0.32\textwidth}
            \includegraphics[width=\textwidth]{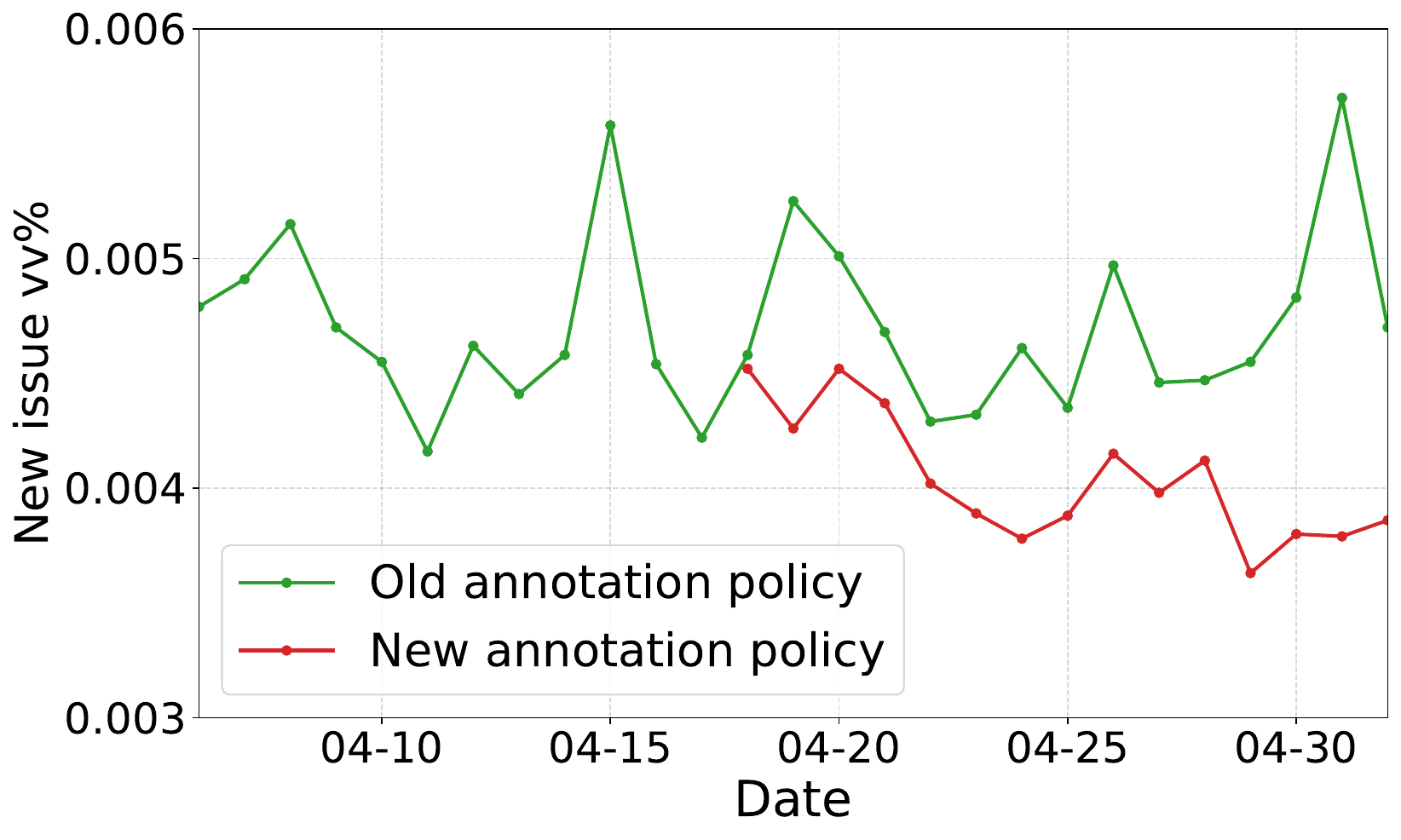}
            \caption{A/B test on old and new versions of annotation policies.}
            \label{fig:ab_test}
        \end{minipage}
        \hfill
        \begin{minipage}[t]{0.31\textwidth}
            \includegraphics[width=\textwidth]{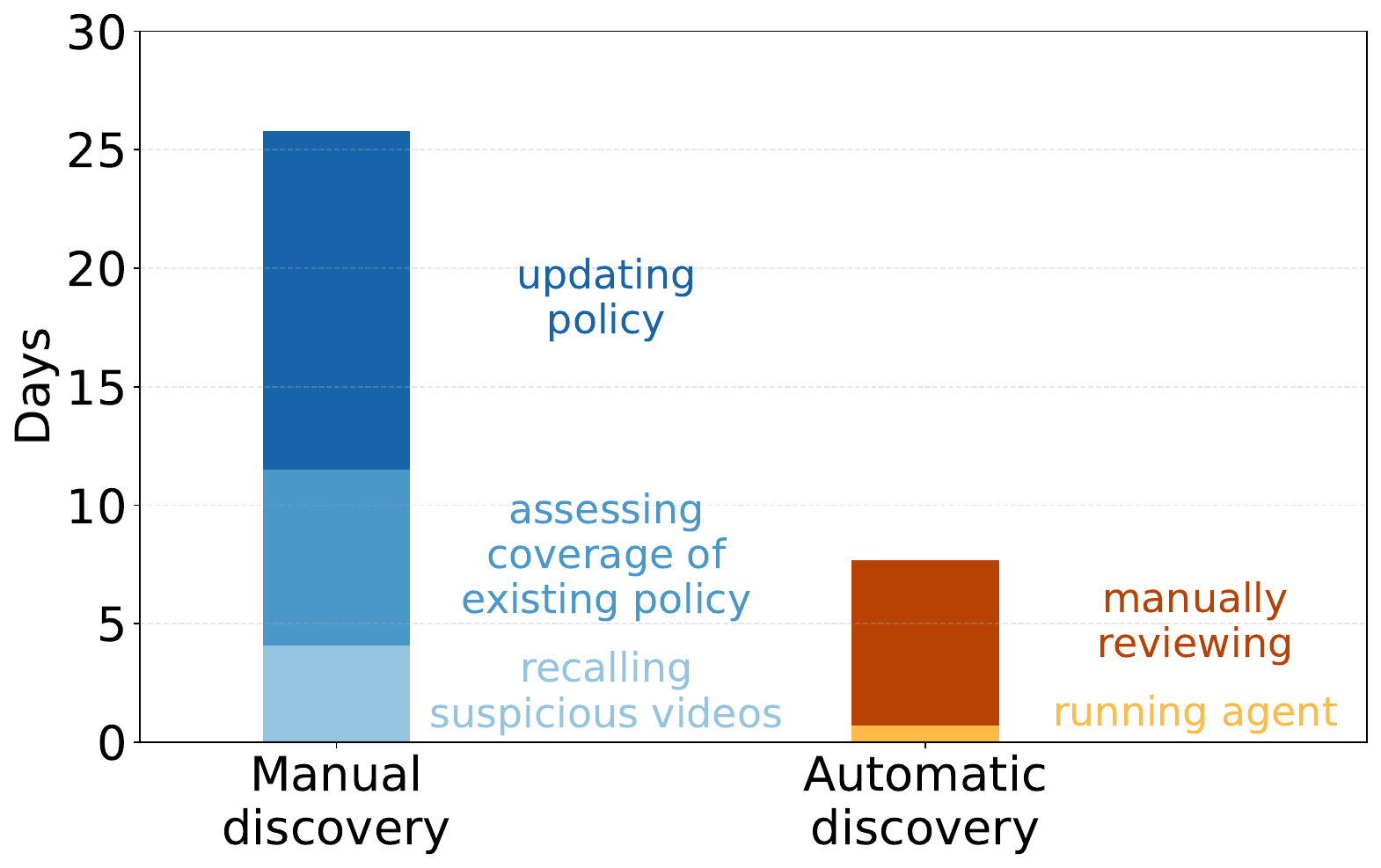}
            \caption{Time cost comparison between manual and automatic issue discovery.}
            \label{fig:efficiency}
        \end{minipage}
    \end{figure*}

    \subsection{Main Results}
        \subsubsection{Offline Comparison with Baseline Methods}
        \label{sec:offline}
        The results of our agent as well as baseline methods on the four-class classification task are presented in Table \ref{tab:main}.
        We have several observations:
        \begin{itemize}
            \item
                Despite not using any training data, the agent consistently outperforms the SFT models.
                Specifically, for CB and UDC, the best-performing agent achieves absolute improvements in macro-F1 of 26.26\% and 22.12\%, respectively, compared with the best-performing SFT model.
            \item
                Comparing \textit{No tools} setting with the \textit{Tool 1} setting (or \textit{Tool 2} vs. \textit{Tools 1\&2}, or \textit{Tool 3} vs. \textit{Tools 1\&3}), we observe that Tool 1 provides a significant performance boost for the agent.
            \item
                Comparing \textit{Tool 2} with \textit{Tool 3} (or \textit{Tools 1\&2} with \textit{Tools 1\&3}), Tool 2 achieves better performance than Tool 3.
                However, note that Tool 2 relies on an embedding model, which introduces additional computation and storage costs.
            \item
                We also examine how the number of training samples for Cases 3/4 affects the SFT model's performance.
                In the CB issue, we gradually increase the number of training samples for Cases 3/4 from 8 to 512 and report the results in Figure \ref{fig:sft}.
                As the number of training samples increases, the SFT's performance improves accordingly.
                Nevertheless, even with 512 training samples (which is nearly impossible in real-world settings due to the scarcity of labeled data containing new issues), the SFT still performs far below the agent.
        \end{itemize}

        \subsubsection{Online A/B Test}
        \label{sec:ab_test}
        As described in Section \ref{sec:deployment}, we design an online A/B experiment to evaluate the performance changes of governance models.
        Our evaluation metric is the ratio of views on videos with new issues to the views of all sampled videos (denoted as \textit{new issue vv\%}).
        A lower ratio indicates that the governance model is more effective in detecting problematic videos and suppressing their distribution within the recommendation system.
        
        Specifically, for CB issue, we apply both the old and new versions of the annotation policy to label the same training dataset and then train two separate governance models.
        We then allocate identical traffic to both models for online testing.
        The old governance model is deployed starting April 6, 2025, while the new model is deployed on April 18, 2025. 
        From these, we sample a subset of videos, manually annotate those with new issues, and calculate new issue vv\%.
        The experimental results, shown in Figure \ref{fig:ab_test}, demonstrate that compared with the old policy, the governance model trained under the new policy significantly reduced new issue vv\%, from an average of 0.0047\% to around 0.0040\%, a decrease of 14.9\%.

        \subsubsection{Efficiency Improvements}
        \label{sec:efficiency}
        For the traditional manual discovery process, we analyze several past cases where the annotation policy was updated and calculate the average time spent at each phase.
        First, the issue discovery team identifies a suspicious video, which may be a variant of an existing issue or a completely new issue.
        On average, this step takes 4.1 days.
        Next, they need to contact the team responsible for defining and maintaining the annotation policy and hold a joint meeting to determine whether the case can be covered by the existing policy.
        This step takes an average of 7.4 days.
        Finally, updating existing sub-issues within the policy or adding new ones requires about 14.3 days on average.
        
        In contrast, for the agent-based process, the discovery step typically takes less than a day, with an average of 0.7 days.
        The generated annotations then undergo human review, which averages 7.0 days.
        This is shorter than the 14.3 days required for manual policy updates, since the policy has already been generated automatically.
        The result of the comparison is shown in Figure \ref{fig:efficiency}.
        Compared with manual discovery that takes an average of 25.8 days, the automated process only requires 7.7 days, reducing the time cost by 70.2\%.

        \subsubsection{Computational Resource Cost}
        We compare the computational resource costs of agent and SFT during deployment.
        On average, we process about 30k videos per day.
        The SFT model achieves a QPS of 5 on H100 GPUs, which means it requires approximately 1.7 H100 GPU-hours per day.
        The agent, on the other hand, generates around five times more queries per video, and each query contains about 1.5 times more tokens.
        As a result, the agent requires about 12.5 H100 GPU-hours per day.
        Although the agent demands more computational resources, the additional 12.5 GPU-hours per day do not impose any noticeable strain on our deployment.

    \subsection{Phase-Wise Results}
        We present the experimental results for each phase.
        The input for each phase is clean (that is, the input for Phase 2 includes only data from Cases 2/3/4, while Phase 3 includes only data from Cases 3/4) to ensure that the results accurately reflect the performance of each phase, without being affected by error propagation from previous phases.
        The results are summarized in Table \ref{tab:phase}.

        \subsubsection{Results of Phase 1: Recalling Suspicious Videos}
        By comparing agent and SFT in Phase 1, we find that the agent performs better overall.
        However, the improvement is less significant than in later phases.
        This may be due to two reasons:
        (1) The differences between normal videos and issue-containing videos are relatively distinct, making the Phase 1 task easier, and thus, the SFT model can also perform quite well;
        (2) The SFT training data contains a large number of samples from Cases 1 and 2, making it relatively easier for the model to distinguish Case 1 from Cases 2/3/4.

        \begin{table}[t]
            \vspace{0.5em}
            \centering
            \setlength{\tabcolsep}{3.5pt}
            \small
            \begin{tabular}{c|cccc|cccc}
                \toprule
                \textbf{Phases \&} & \multicolumn{4}{c|}{\textbf{CB}} & \multicolumn{4}{c}{\textbf{UDC}} \\
                \textbf{Methods} & \textit{P} & \textit{R} & \textit{M-F1} & \textit{ARI} & \textit{P} & \textit{R} & \textit{M-F1} & \textit{ARI} \\
                \midrule
                \midrule
                %\multicolumn{1}{c}{\textbf{Issue discovery agent}} & \multicolumn{4}{c}{} \\
                \multicolumn{1}{c}{\textbf{Phase 1}} & \multicolumn{8}{c}{} \\
                \makebox[4.5em][l]{\textit{w/ agent}} & 79.50 & \textbf{89.03} & \textbf{83.34} & - & 78.97 & \textbf{81.22} & \textbf{80.03} & - \\
                \makebox[4.5em][l]{\textit{w/ SFT}} & \textbf{82.76} & 78.71 & 80.57 & - & \textbf{92.74} & 59.81 & 64.49 & - \\
                \midrule
                \midrule
                \multicolumn{1}{c}{\textbf{Phase 2}} & \multicolumn{8}{c}{} \\
                \makebox[4.5em][l]{\textit{w/ agent}} & 63.45 & 51.79 & 33.58 & - & 70.83 & 61.45 & 65.81 & - \\
                \makebox[4.5em][l]{\quad + \textit{Tool 1}} & \textbf{87.10} & \textbf{63.78} & \textbf{73.64} & - & \textbf{71.64} & \textbf{71.57} & \textbf{71.60} & - \\
                \makebox[4.5em][l]{\textit{w/ SFT}} & 34.37 & 47.09 & 36.97 & - & 30.08 & 50.10 & 37.56 & - \\
                \midrule
                \midrule
                \multicolumn{1}{c}{\textbf{Phase 3}} & \multicolumn{8}{c}{} \\
                \makebox[4.5em][l]{\textit{w/ agent}} & 40.87 & 49.52 & 44.78 & - & 37.95 & 50.00 & 43.15 & - \\
                \makebox[4.5em][l]{\quad + \textit{Tool 2}} & 51.35 & \textbf{73.08} & \textbf{60.32} & \textbf{0.39} & \textbf{93.27} & 73.08 & \textbf{77.97} & \textbf{0.47} \\
                \makebox[4.5em][l]{\quad + \textit{Tool 3}} & \textbf{57.61} & 54.85 & 55.18 & 0.31 & 80.05 & \textbf{74.33} & 76.47 & 0.40 \\
                \makebox[4.5em][l]{\textit{w/ SFT}} & 39.92 & 50.00 & 44.40 & - & 36.79 & 50.00 & 42.39 & - \\
                \bottomrule
            \end{tabular}
            \vspace{1ex}
            \caption{Phase-wise result of agent and SFT. The highest value in each column of each block is shown in bold. P: Precision; R: Recall; M-F1: Macro-F1; ARI: Adjusted Rand Index.}
            \vspace{-2ex}
            \label{tab:phase}
        \end{table}

        \subsubsection{Results of Phase 2: Assessing Coverage of Existing Policies}
        From the Phase 2 results, we can see that the agent significantly outperforms SFT, and the performance gain brought by Tool 1 is particularly significant.
        This indicates that the agent is capable of autonomously invoking governance models to enhance its decision accuracy effectively.

        \subsubsection{Results of Phase 3: Determining Variants vs. New Issues}
        In Phase 3, when the agent operates without any tools and relies solely on its own judgment, its performance is relatively poor, comparable to that of the SFT model.
        Moreover, without tools, the agent can only perform Case 3 vs. Case 4 classification, but cannot conduct sub-issue-level clustering.
        
        After incorporating tools, the agent's performance improves substantially.
        As discussed in Section \ref{sec:offline}, Tool 2 performs better than Tool 3.
        This advantage is reflected not only in Case 3 vs. Case 4 classification accuracy, but also in clustering performance.
        We evaluate clustering quality using the Adjusted Rand Index (ARI), a metric that measures the consistency between clustering results and ground-truth labels.
        The ARI ranges from (-1, 1], where a value of 1 indicates perfect agreement with the true labels, 0 denotes completely random clustering, and -1 indicates complete disagreement.
        The ARI results for Tool 2 and Tool 3 further confirm that Tool 2 achieves better clustering performance.

    \begin{figure}[t]
        \setlength{\belowcaptionskip}{-3pt}
        \centering
        \includegraphics[width=\linewidth]{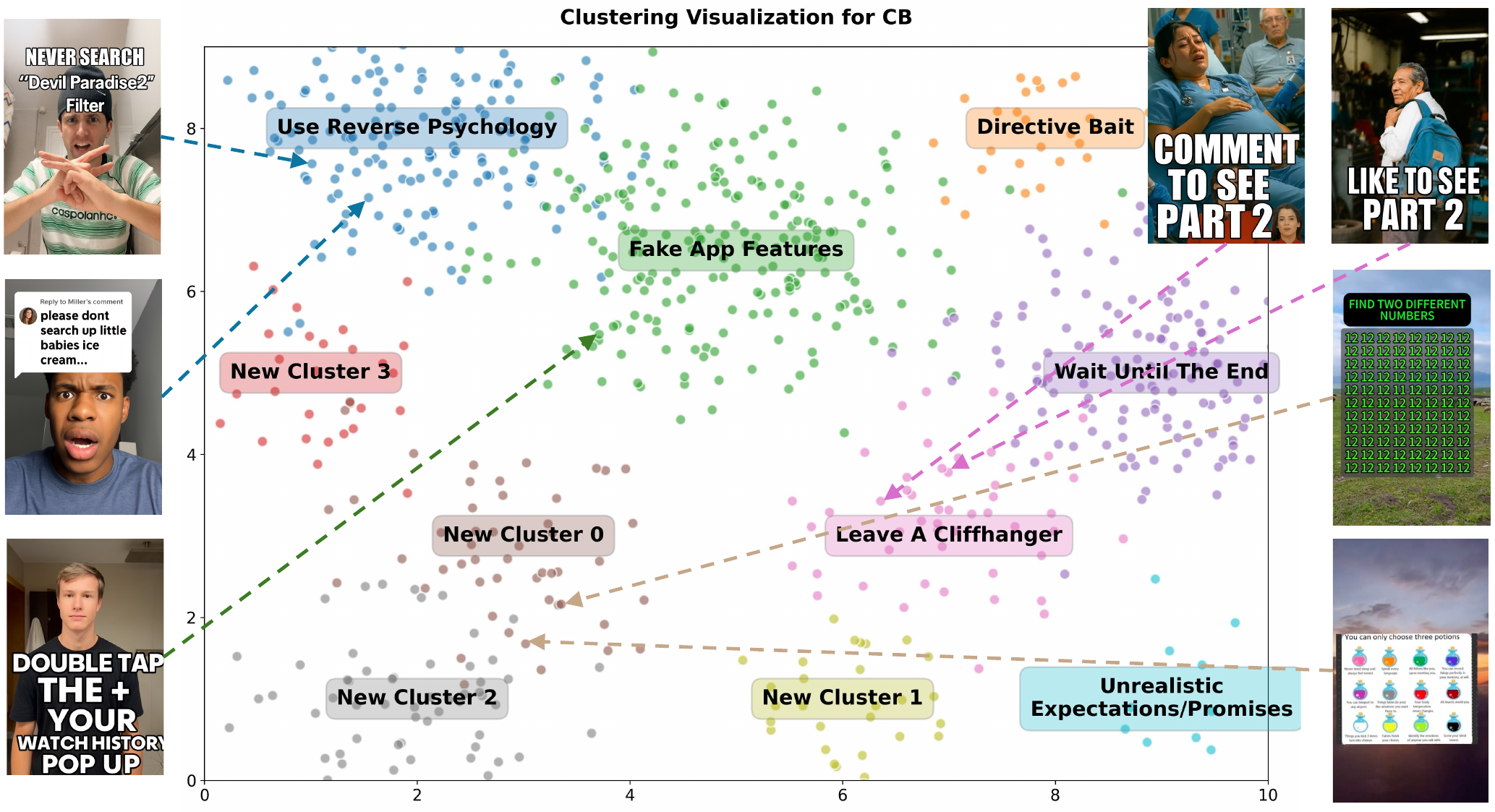}
        \caption{Visualized clustering result in CB.}
        \label{fig:case_study_clustering}
    \end{figure}

    \begin{figure}[t]
        \setlength{\belowcaptionskip}{-3pt}
        \centering
        \includegraphics[width=0.96\linewidth]{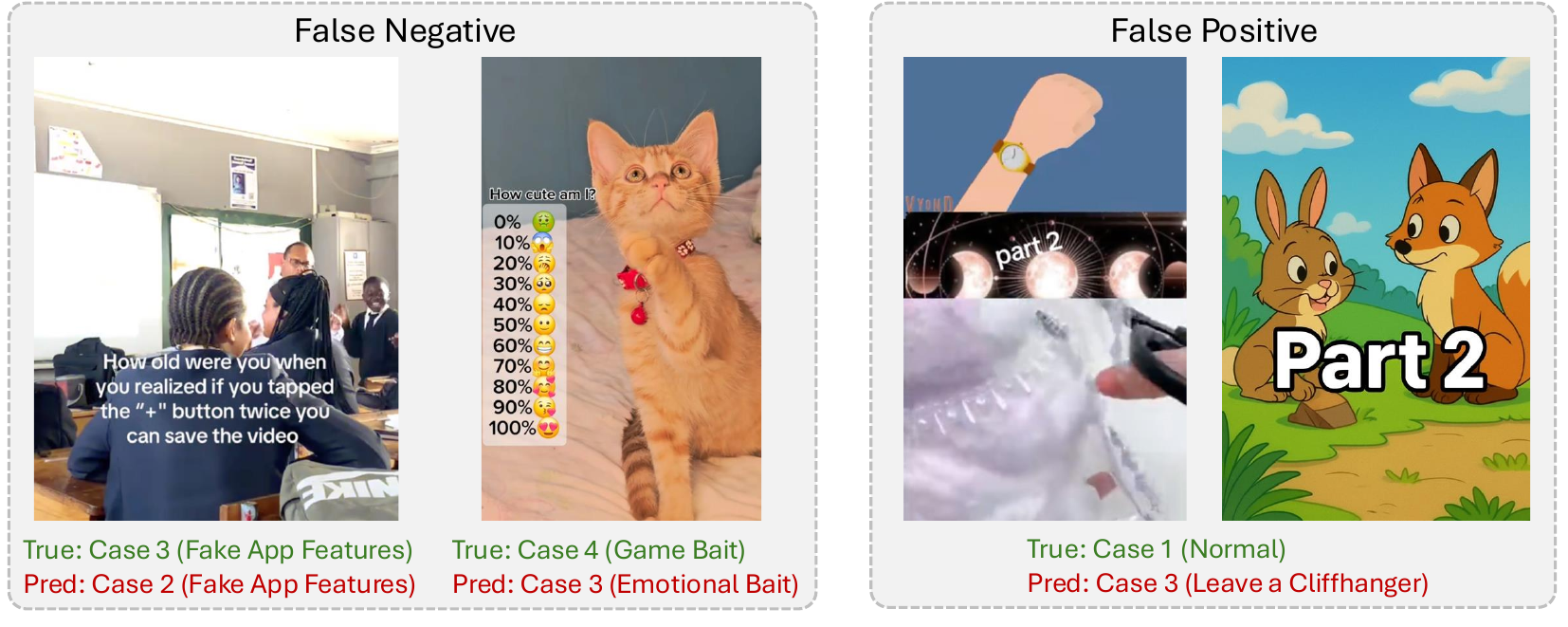}
        \caption{Examples of failure cases in CB.}
        \label{fig:failure_case}
    \end{figure}
    
    \subsection{Success \& Failure Case Analysis}
        \subsubsection{Video Clustering}
        We visualize the clustering results in the CB using t-SNE \cite{maaten2008visualizing}.
        As shown in Figure \ref{fig:case_study_clustering}, the agent successfully identifies many existing videos that contain variants of existing sub-issues, such as \textsf{Use reverse psychology}, \textsf{Fake app features}, and \textsf{Leave a cliffhanger}.
        Moreover, within the clusters of Case-4 videos, we discovered that \textsf{New Cluster 0} represents a typical new sub-issue, \textsf{Game bait}, where content is presented in the form of games or interactive choices to engage users.

        Figure \ref{fig:failure_case} presents several failure cases from the agent's predictions.
        In the first example, the agent incorrectly classifies the video as being covered by existing policies, possibly because the phrase ``save the video'' appears in that policy.
        The second example is misclassified due to its similarity to \textsf{Emotional bait} (which appeals to morality, gratitude, or sympathy in a coercive way to pressure viewers into engaging).
        The last two examples are actually normal videos, but the agent mistakenly identifies them as \textsf{Leave a cliffhanger} due to the presence of ``Part 2'' in the content (see related examples in Figure \ref{fig:case_study_clustering}).
        Overall, these are challenging borderline cases where the agent tends to make more mistakes, and we plan to include more user-interaction-related features to assist the agent in making judgments, such as ratings, shares, and comments.

        \subsubsection{Generated Annotation Policies}
\begin{figure}[t]
        \setlength{\belowcaptionskip}{-3pt}
        \centering
        \includegraphics[width=0.98\linewidth]{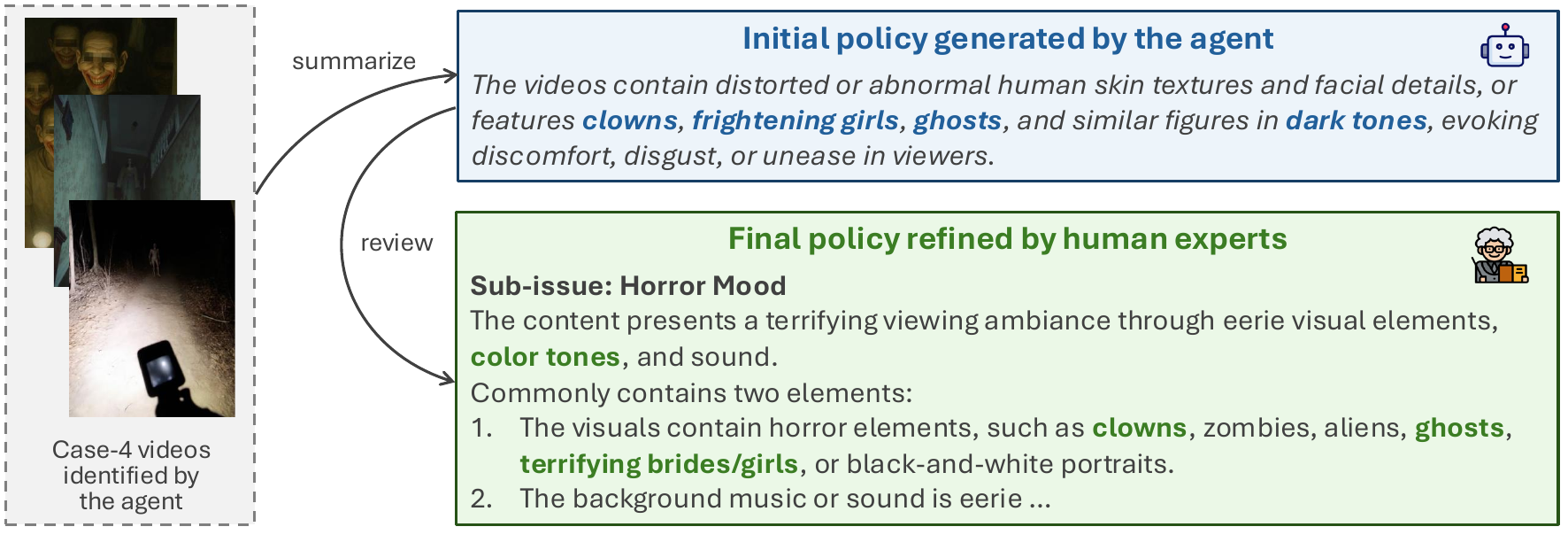}
        \caption{Examples of policy generation in UDC.}
        \label{fig:generated_policy}
    \end{figure}

        We illustrate the policy generated by the agent for a new sub-issue in UDC.
        As shown in Figure \ref{fig:generated_policy}, the agent identifies a group of Case-4 videos characterized by horror elements set in dark scenes.
        It then summarizes the cluster and proposes the following sub-issue definition: ``\textsf{... or features clowns, frightening girls, ghosts, and similar figures in dark tones.}''
        This new issue definition is subsequently reviewed by human experts, and the final revised annotation policy is: ``\textsf{The content presents a terrifying viewing ambiance through eerie visual elements and color tones ... such as clowns, ..., ghosts, terrifying brides/girls.}''
        As we can see, much of the content generated by the agent is eventually adopted in the final policy (the highlighted parts in Figure \ref{fig:generated_policy}).
        This significantly improves efficiency compared to the traditional fully manual process of creating and updating policies.

\section{Conclusion and Future Work}
    The paper introduces our progress in using agent-based LLMs for emerging issue discovery on short-video platforms.
    The agent operates through four phases, which first progressively filters out short videos containing new issues, then applies a streaming clustering method to further group them into variants of existing sub-issues or new sub-issues.
    Finally, the agent generates an updated annotation policy.
    Compared with manual discovery of new issues, our method significantly improves efficiency and reduces time costs.
    
    We plan to integrate more advanced tools into this agent framework.
    For example, \textit{intelligent video frame sampling}: currently, video frames are uniformly sampled, which often leads to redundant frames while missing fleeting but important details.
    By adopting smarter sampling, the agent can extract frames most likely to contain issues and filter out irrelevant ones. Another direction is \textit{external search}: by accessing the Internet, the agent can incorporate up-to-date information on current events and trends, allowing it to make more accurate judgments about emerging issues.
    
    Looking back at the overall deployment scheme in Figure \ref{fig:deployment}, it is clear that video annotation still relies entirely on manual labor, which remains highly time-consuming and costly.
    Exploring ways to bring more automation into the annotation process is, therefore, another important direction for our future research.

\bibliographystyle{ACM-Reference-Format}
\bibliography{reference}

@misc{tiktok_guidelines_2025,
  author       = {TikTok Inc.},
  title        = {TikTok Community Guidelines},
  year         = {2025},
  howpublished = {\url{https://www.tiktok.com/community-guidelines/en}}
}

@article{zubarouglu2021data,
  title={Data stream clustering: a review},
  author={Zubaro{\u{g}}lu, Alaettin and Atalay, Volkan},
  journal={Artificial Intelligence Review},
  volume={54},
  number={2},
  pages={1201--1236},
  year={2021},
  publisher={Springer}
}

@inproceedings{o2002streaming,
  title={Streaming-data algorithms for high-quality clustering},
  author={O'callaghan, Liadan and Mishra, Nina and Meyerson, Adam and Guha, Sudipto and Motwani, Rajeev},
  booktitle={Proceedings 18th international conference on data engineering},
  pages={685--694},
  year={2002},
  organization={IEEE}
}

@inproceedings{radford2021learning,
  title={Learning transferable visual models from natural language supervision},
  author={Radford, Alec and Kim, Jong Wook and Hallacy, Chris and Ramesh, Aditya and Goh, Gabriel and Agarwal, Sandhini and Sastry, Girish and Askell, Amanda and Mishkin, Pamela and Clark, Jack and others},
  booktitle={International conference on machine learning},
  pages={8748--8763},
  year={2021},
  organization={PmLR}
}

@inproceedings{wulongmemeval,
  title={LongMemEval: Benchmarking Chat Assistants on Long-Term Interactive Memory},
  author={Wu, Di and Wang, Hongwei and Yu, Wenhao and Zhang, Yuwei and Chang, Kai-Wei and Yu, Dong},
  booktitle={The Thirteenth International Conference on Learning Representations}
}

@article{wang2023augmenting,
  title={Augmenting language models with long-term memory},
  author={Wang, Weizhi and Dong, Li and Cheng, Hao and Liu, Xiaodong and Yan, Xifeng and Gao, Jianfeng and Wei, Furu},
  journal={Advances in Neural Information Processing Systems},
  volume={36},
  pages={74530--74543},
  year={2023}
}

@inproceedings{bhagoji2024community,
  author    = {Bhagoji, A. and Schaffner, B. and others},
  title     = {{'Community Guidelines Make This the Best Party on the Internet'}: An In-Depth Study of Online Platforms' Content Moderation Policies},
  booktitle = {Proceedings of the 2024 CHI Conference on Human Factors in Computing Systems},
  year      = {2024},
  address   = {New York, NY, USA},
  pages     = {1--15},
  doi       = {10.1145/3613904.3642345},
  publisher = {ACM}
}

@misc{youtube2023guidelines,
  author       = {YouTube},
  title        = {YouTube Community Guidelines Enforcement Report},
  year         = {2023},
  howpublished = {\url{https://transparencyreport.google.com/youtube-policy/overview}}
}

@inproceedings{dosovitskiy2021image,
  author    = {Dosovitskiy, A. and Beyer, L. and others},
  title     = {An Image is Worth 16x16 Words: Transformers for Image Recognition at Scale},
  booktitle = {Proceedings of the International Conference on Learning Representations (ICLR)},
  year      = {2021},
  url       = {https://openreview.net/forum?id=YicbFdNTTy}
}

@article{li2020multiscale,
  author  = {Li, X. and Wang, Y. and others},
  title   = {Multi-Scale Context Aggregation for Video Violence Detection},
  journal = {IEEE Transactions on Multimedia},
  year    = {2020},
  volume  = {22},
  number  = {5},
  pages   = {1234--1247},
  month   = {May},
  doi     = {10.1109/TMM.2019.2940099}
}

@misc{facebook2022aisafety,
  author       = {Facebook AI},
  title        = {Advancing AI for Content Safety: Updates and Insights},
  year         = {2022},
  howpublished = {\url{https://ai.facebook.com/research/advancing-ai-for-content-safety/}},
}

@misc{hosseini2023faster,
  author        = {Hosseini, M. A. and Hasan, M.},
  title         = {Faster, Lighter, More Accurate: A Deep Learning Ensemble for Content Moderation},
  year          = {2023},
  eprint        = {2309.05150},
  archiveprefix = {arXiv}
}

@misc{ahmed2024enhanced,
  author        = {Ahmed, S. H. and Khan, M. J. and Sukthankar, G.},
  title         = {Enhanced Multimodal Content Moderation of Children's Videos using Audiovisual Fusion},
  year          = {2024},
  eprint        = {2405.06128},
  archiveprefix = {arXiv},
  primaryclass  = {cs.CV}
}

@misc{aldahoul2024advancing,
  author        = {Al Dahoul, N. and Toledo Tan, M. J. and others},
  title         = {Advancing Content Moderation: Evaluating Large Language Models for Detecting Sensitive Content Across Text, Images, and Videos},
  year          = {2024},
  eprint        = {2411.17123},
  archiveprefix = {arXiv}
}

@inproceedings{wang-etal-2025-filter,
    title      = "Filter-And-Refine: A {MLLM} Based Cascade System for Industrial-Scale Video Content Moderation",
    author     = "Wang, Zixuan and Shi, Jinghao  and Liang, Hanzhong  and Shen, Xiang  and Wen, Vera  and Chen, Zhiqian  and Wu, Yifan  and Zhang, Zhixin  and Xiong, Hongyu",
    publisher  = "Association for Computational Linguistics",
    url        = "https://aclanthology.org/2025.acl-industry.62/",
}

@misc{oak2025reranking,
  author        = {Oak, R. and Haroon, M. and others},
  title         = {Re-ranking Using Large Language Models for Mitigating Exposure to Harmful Content on Social Media Platforms},
  year          = {2025},
  eprint        = {2501.13977},
  archiveprefix = {arXiv}
}

@inproceedings{han2023traditional,
  author    = {Han, Jiawei and Kamber, Micheline},
  title     = {Advances and Limitations of Traditional Data Mining for Knowledge Discovery},
  booktitle = {Proc. 2023 IEEE Int. Conf. Data Mining (ICDM)},
  year      = {2023},
  pages     = {1--8},
  doi       = {10.1109/ICDM54844.2023.00002}
}

@article{agrawal1994fast,
  author  = {Agrawal, Rakesh and Srikant, Ramakrishnan},
  title   = {Fast Algorithms for Mining Association Rules in Large Databases},
  journal = {Proc. 20th Int. Conf. Very Large Data Bases (VLDB)},
  year    = {1994},
  pages   = {487--499}
}

@article{zhang2024llm,
  author  = {Zhang, Lijun and Chen, Wei and Liu, Zhiyuan},
  title   = {Prompt-Tuned LLMs for Low-Resource Domain Knowledge Discovery},
  journal = {IEEE Trans. Knowl. Data Eng.},
  year    = {2024},
  volume  = {36},
  number  = {5},
  pages   = {2189--2203},
  doi     = {10.1109/TKDE.2023.3326457}
}

@inproceedings{rajagopal2023llm,
  author    = {Rajagopal, Ram and Srinivasan, Balaji and Ethayarajh, Kawin},
  title     = {LLM-Driven Knowledge Graph Construction from Scientific Literature},
  booktitle = {Proc. 2023 ACM SIGKDD Conf. Knowl. Discovery Data Mining (KDD)},
  year      = {2023},
  pages     = {2045--2055},
  doi       = {10.1145/3580305.3599256}
}

@inproceedings{wang2023multi,
  author    = {Wang, Hao and Li, Jiong and Chen, Xi},
  title     = {Cross-Modal Attention for Multi-Modal Knowledge Discovery in Medical Imaging},
  booktitle = {Proc. 2023 IEEE Conf. Comput. Vis. Pattern Recognit. (CVPR)},
  year      = {2023},
  pages     = {14567--14576},
  doi       = {10.1109/CVPR52729.2023.01402}
}

@article{choi2024healthcare,
  author  = {Choi, Edward and Kim, Sungmin and Shin, Hwanhee},
  title   = {Knowledge Discovery from EHRs for Personalized Treatment Recommendation},
  journal = {NPJ Digit. Med.},
  year    = {2024},
  volume  = {7},
  number  = {1},
  pages   = {45},
  doi     = {10.1038/s41746-024-01189-x}
}

@inproceedings{jiang2023finance,
  author    = {Jiang, Tao and Yu, Han and Zhang, Wei},
  title     = {Knowledge Discovery in Financial Market Data for Algorithmic Trading Risk Assessment},
  booktitle = {Proc. 2023 Int. Conf. Financ. Cryptogr. Data Secur. (FC)},
  year      = {2023},
  pages     = {345--362},
  doi       = {10.1007/978-3-031-37993-3_18}
}

@article{wooldridge2022foundations,
  author  = {Wooldridge, Michael},
  title   = {Foundations of Intelligent Agents},
  journal = {Annual Review of Control, Robotics, and Autonomous Systems},
  year    = {2022},
  volume  = {5},
  pages   = {1--25},
  doi     = {10.1146/annurev-control-042921-020815}
}

@article{wei2022chain,
  author  = {Wei, Jason and Wang, Xuezhi and Schuurmans, Dale and Bosma, Maarten and Ichter, Brian and Xia, Fei and Zhou, Denny and Le, Quoc V and Fedus, William},
  title   = {Chain-of-Thought Prompting Elicits Reasoning in Large Language Models},
  journal = {Advances in Neural Information Processing Systems},
  year    = {2022},
  volume  = {35},
  pages   = {24824--24837}
}

@inproceedings{yao2023tree,
  author    = {Yao, Shunyu and Yu, Dian and Zhao, Jeffrey and Shafran, Izhak and Griffiths, Thomas L and Cao, Yuan and Narasimhan, Karthik},
  title     = {Tree-of-Thoughts: Deliberate Problem Solving with Large Language Models},
  booktitle = {Proceedings of the 37th Conference on Neural Information Processing Systems},
  year      = {2023},
  pages     = {1--14}
}

@inproceedings{yao2022react,
  author    = {Yao, Shunyu and Zhao, Jeffrey},
  title     = {ReAct: Synergizing Reasoning and Acting in Language Models},
  booktitle = {Proceedings of the 36th Conference on Neural Information Processing Systems},
  year      = {2022},
  pages     = {1--12}
}

@article{schuurmans2023toolformer,
  author  = {Schuurmans, Dale and Bosma, Maarten and Ichter, Brian and Xia, Fei and Zhou, Denny and Fedus, William and Xu, Elvis and Kordi, Yikang and Mishra, Shantanu and Yu, Athul and others},
  title   = {Toolformer: Language Models Can Teach Themselves to Use Tools},
  journal = {Advances in Neural Information Processing Systems},
  year    = {2023},
  volume  = {36},
  pages   = {21480--21494}
}

@article{yu2024usm,
  title={USM: Unbiased Survey Modeling for Limiting Negative User Experiences in Recommendation Systems},
  author={Yu, Chenghui and Li, Peiyi and Wu, Haoze and Wen, Yiri and Deng, Bingfeng and Xiong, Hongyu},
  journal={arXiv preprint arXiv:2412.10674},
  year={2024}
}

@article{zhang2024optimization,
  author  = {Zhang, Han and Li, Zhen and Wang, Jianmin and Tang, Jie},
  title   = {A Survey on Optimization Methods for LLM-Based Agents},
  journal = {IEEE Transactions on Knowledge and Data Engineering},
  year    = {2024},
  volume  = {36},
  number  = {8},
  pages   = {3845--3862},
  doi     = {10.1109/TKDE.2024.3387645}
}

@article{nakano2021webgpt,
  title={Webgpt: Browser-assisted question-answering with human feedback},
  author={Nakano, Reiichiro and Hilton, Jacob and Balaji, Suchir and Wu, Jeff and Ouyang, Long and Kim, Christina and Hesse, Christopher and Jain, Shantanu and Kosaraju, Vineet and Saunders, William and others},
  journal={arXiv preprint arXiv:2112.09332},
  year={2021}
}

@article{shen2023hugginggpt,
  title={Hugginggpt: Solving ai tasks with chatgpt and its friends in hugging face},
  author={Shen, Yongliang and Song, Kaitao and Tan, Xu and Li, Dongsheng and Lu, Weiming and Zhuang, Yueting},
  journal={Advances in Neural Information Processing Systems},
  volume={36},
  pages={38154--38180},
  year={2023}
}

@article{hu2022lora,
  title={Lora: Low-rank adaptation of large language models.},
  author={Hu, Edward J and Shen, Yelong and Wallis, Phillip and Allen-Zhu, Zeyuan and Li, Yuanzhi and Wang, Shean and Wang, Lu and Chen, Weizhu and others},
  journal={ICLR},
  volume={1},
  number={2},
  pages={3},
  year={2022}
}

@article{bai2025qwen2,
  title={Qwen2. 5-vl technical report},
  author={Bai, Shuai and Chen, Keqin and Liu, Xuejing and Wang, Jialin and Ge, Wenbin and Song, Sibo and Dang, Kai and Wang, Peng and Wang, Shijie and Tang, Jun and others},
  journal={arXiv preprint arXiv:2502.13923},
  year={2025}
}

@article{bao2022vlmo,
  title={Vlmo: Unified vision-language pre-training with mixture-of-modality-experts},
  author={Bao, Hangbo and Wang, Wenhui and Dong, Li and Liu, Qiang and Mohammed, Owais Khan and Aggarwal, Kriti and Som, Subhojit and Piao, Songhao and Wei, Furu},
  journal={Advances in neural information processing systems},
  volume={35},
  pages={32897--32912},
  year={2022}
}

@inproceedings{cherti2023reproducible,
  title={Reproducible scaling laws for contrastive language-image learning},
  author={Cherti, Mehdi and Beaumont, Romain and Wightman, Ross and Wortsman, Mitchell and Ilharco, Gabriel and Gordon, Cade and Schuhmann, Christoph and Schmidt, Ludwig and Jitsev, Jenia},
  booktitle={Proceedings of the IEEE/CVF conference on computer vision and pattern recognition},
  pages={2818--2829},
  year={2023}
}

@article{liu2023visual,
  title={Visual instruction tuning},
  author={Liu, Haotian and Li, Chunyuan and Wu, Qingyang and Lee, Yong Jae},
  journal={Advances in neural information processing systems},
  volume={36},
  pages={34892--34916},
  year={2023}
}

@article{ahmed2020k,
  title={The k-means algorithm: A comprehensive survey and performance evaluation},
  author={Ahmed, Mohiuddin and Seraj, Raihan and Islam, Syed Mohammed Shamsul},
  journal={Electronics},
  volume={9},
  number={8},
  pages={1295},
  year={2020},
  publisher={MDPI}
}

@article{murtagh2012algorithms,
  title={Algorithms for hierarchical clustering: an overview},
  author={Murtagh, Fionn and Contreras, Pedro},
  journal={Wiley interdisciplinary reviews: data mining and knowledge discovery},
  volume={2},
  number={1},
  pages={86--97},
  year={2012},
  publisher={Wiley Online Library}
}

@inproceedings{yu2025unified,
  title={Unified Survey Modeling to Limit Negative User Experiences in Recommendation Systems},
  author={Yu, Chenghui and Wu, Haoze and Ding, Jian and Deng, Bingfeng and Xiong, Hongyu},
  booktitle={Proceedings of the Nineteenth ACM Conference on Recommender Systems},
  pages={1104--1107},
  year={2025}
}

@article{wang2025reasoning,
  title={Reasoning-Enhanced Domain-Adaptive Pretraining of Multimodal Large Language Models for Short Video Content Governance},
  author={Wang, Zixuan and Sun, Yu and Wang, Hongwei and Jing, Baoyu and Shen, Xiang and Dong, Xin and Hao, Zhuolin and Xiong, Hongyu and Song, Yang},
  journal={arXiv preprint arXiv:2509.21486},
  year={2025}
}

@article{maaten2008visualizing,
  title={Visualizing data using t-SNE},
  author={Maaten, Laurens van der and Hinton, Geoffrey},
  journal={Journal of machine learning research},
  volume={9},
  number={Nov},
  pages={2579--2605},
  year={2008}
}

\end{document}